\documentclass[10pt,twocolumn,letterpaper]{article}

\usepackage{wacv}
\usepackage{times}
\usepackage{epsfig}
\usepackage{graphicx}
\usepackage{amsmath}
\usepackage{amssymb}
\usepackage{subfig}
\usepackage{booktabs}
\usepackage{multirow}
\usepackage{url}
\usepackage{dirtytalk}

\newcommand{\method}{C4Synth\xspace}

\newcommand{\first}{Cascaded-C4Synth\xspace}
\newcommand{\second}{Recurrent-C4Synth\xspace}
\newcommand{\ccc}{Cross-Caption Cycle Consistency\xspace} 

\wacvfinalcopy 

\def\wacvPaperID{271} 

\ifwacvfinal\fi
\begin{document}

\title{C4Synth: Cross-Caption Cycle-Consistent Text-to-Image Synthesis}

\author{K J Joseph \hspace{0.5cm} Arghya Pal \hspace{0.5cm} Sailaja Rajanala \hspace{0.5cm} Vineeth N Balasubramanian\\
IIT Hyderabad, India\\
{\tt\small cs17m18p100001@iith.ac.in}
}

\maketitle
\ifwacvfinal\thispagestyle{empty}\fi

\begin{abstract}
Generating an image from its description is a challenging task worth solving because of its numerous practical applications ranging from image editing to virtual reality. All existing methods use one single caption to generate a plausible image. A single caption by itself, can be limited, and may not be able to capture the variety of concepts and behavior that may be present in the image. We propose two deep generative models that generate an image by making use of multiple captions describing it. This is achieved by ensuring `\ccc' between the multiple captions and the generated image(s). We report quantitative and qualitative results on the standard Caltech-UCSD Birds (CUB) and Oxford-102 Flowers datasets to validate the efficacy of the proposed approach.
\end{abstract}

\section{Introduction} \label{sec:intro}
The days when imagination was constrained by a human's visualizing capabilities are gradually passing behind us. Through text to image synthesis, works such as
~\cite{reed2016generative, xu2017attngan,Han17stackgan2,han2017stackgan} have introduced us to the possibility of visualizing through textual descriptions. Text-to-image synthesis has found itself a foothold in many real-world applications such as virtual reality, tourism, image editing, gaming, and computer-aided design. More mathematically said, the problem is that of modeling $P(\textbf{I}|\textbf{t})$: $\textbf{I}$ being the generated image, $\textbf{t}$ the raw text (user descriptions). Conditioning $P(\textbf{I}|\textbf{t})$ on raw text may not essentially capture the details since the descriptions themselves could be vague. In general, the current trend to overcome this, is to employ distributed text representations to encode the word as a function of its concepts, yielding a text encoding $\phi(\textbf{t})$. This brings conceptually similar words together and scatters the dissimilar words, giving us a rich text representation to model \vspace{0.2cm} $P(\textbf{I}|\phi(\textbf{t}))$ rather than $P(\textbf{I}|\textbf{t})$.

However, as the saying goes: `A picture is worth a thousand words', the information that is conveyed by the visual perception of an image is difficult to be captured by a single textual description (caption) of the image. In order to alleviate this semantic gap, standard image captioning datasets like COCO \cite{lin2014microsoft} and Pascal Sentences \cite{rashtchian2010collecting} provide five captions per image. We show how the use of multiple captions that contain complementary information aid in generating lucid images. It is analogous to having a painter update a canvas each time, after reading different descriptions of the end image that (s)he is painting. 
Captions with unseen information help the artist to add new concepts to the existing canvas. On the other hand, a caption with redundant concepts improves the rendering of the existing sketch.

\begin{figure}
\centering
{\includegraphics[scale=0.72]{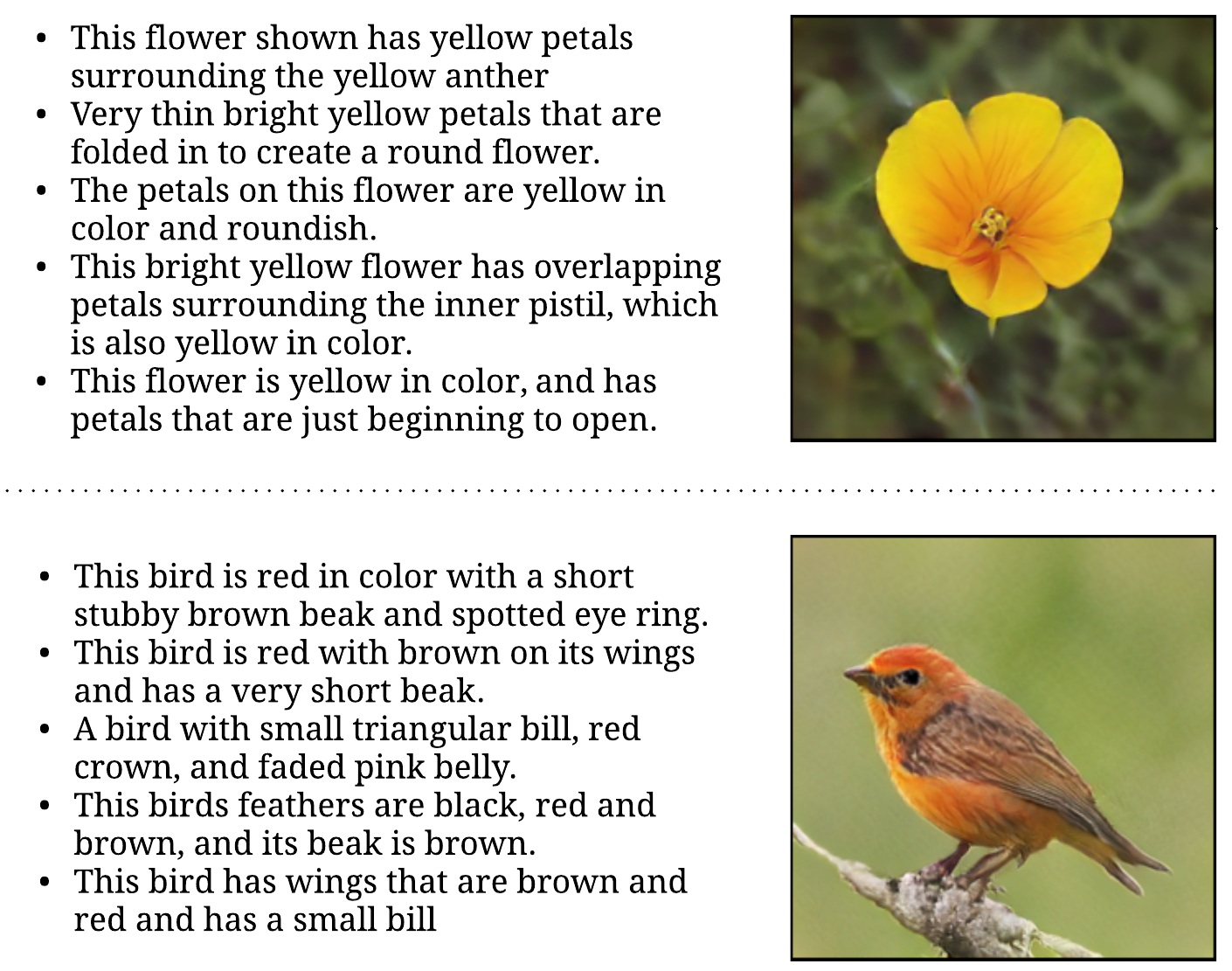}}\qquad
\caption{The figure shows two images generated by \method. The corresponding captions that are used while generating images are listed on the left side. 
(Best viewed in color.)}
\label{image:front_image}
\end{figure}

We realize the aforementioned vision via \method, a deep generative model that iteratively updates its generated image features by taking into account different captions at each step. 
To assimilate information from multiple captions, we use an adversarial learning setup using Generative Adversarial Networks (GANs)~\cite{goodfellow2014generative}, consisting of $i$ generator-discriminator pairs in a serial manner: each conditioned on the current caption encoding $\phi(\textbf{t})_i$ and \textit{history} block $B_i$. We ensure that the captions and the generated image features satisfy a cyclical consistency across the set of captions available. Concretely, let $F_i:\textbf{t}_i\to\textbf{I}_i$ and $G_i:\textbf{I}_i\to\textbf{t}_i$; where $\textbf{t}$ represents a caption, $\textbf{I}$ represents an image, $F_i$ transforms the $i^{th}$ caption to the corresponding image representation and $G_i$ does the opposite. A network that is consistent with two captions, for example, is trained such that $G_2 \circ F_2 \circ G_1 \circ F_1 (\textbf{t}) \approx \textbf{t} $. This model takes inspiration from Cycle-GAN \cite{zhu2017unpaired} which has demonstrated superior performance in unpaired image-to-image translation. We delve into the details of the cycle-consistency and how this is implemented through a cascaded approach, which we call \first, in Section~\ref{sec_methodology}.

The scope of \first is limited by the number of generator-discriminator pairs which are in turn dependent on the number of captions at hand and requires training multiple generator-discriminator pairs in a serial manner. However, the number of available captions can vary across the datasets. 
This calls for a recurrent version of \first, which we christen as \second, which is not bound by the number of captions. 
In \second, the images are generated conditioned on a caption and a hidden-state which acts as a memory to capture dependencies among multiple captions. The architecture is explained in detail in Section \ref{sec:RecurrentMethod}. 

The key contributions of this work are as follows:
\begin{itemize}
\itemsep0em
\item We propose a methodology for image generation using a medley of captions. To the best of our knowledge, this is the first such effort. 
\item We introduce a Cross-Caption Cycle-Consistency (and hence, the name \method) as a means of amalgamating information in multiple concepts to generate a single image. 
\item We supplement the abovementioned method by inducing a recurrent structure that removes the limitation of number of captions on the architecture.
\item Our experiments (both qualitative and quantitative) on the standard Caltech-UCSD Birds (CUB) \cite{WelinderEtal2010} and Oxford-102 Flowers \cite{Nilsback08} datasets show that both our models, \first and \second, generate real-like, plausible images given a set of captions per sample. As an interesting byproduct, we showcase the model's capability to generate stylized images that vary in the pose and background, however, consistent with the set of captions.
\end{itemize}

\section{Related Work} 
\label{sec:relatedWork}
\textbf{Text to Image Synthesis:} In the realm of GANs, text-to-image synthesis is qualified as a conditional Generative Adversarial Network (cGAN) that transforms human-written text descriptions to the pixel space. The admission of a text into pixel space was realized using deep symmetric structured joint embeddings followed by a cGAN in Reed's \etal seminal work \cite{reed2016generative}. This was the first end-to-end differentiable architecture from character-level to pixel-level generation and showed efficacy in generating real-like images. Subsequently, StackGAN \cite{han2017stackgan} and its follow-up work, StackGAN++ \cite{Han17stackgan2}, increased the spatial resolution of the generated image by adopting a two-stage process. In StackGAN, low-resolution images ($64 \times 64$) generated by the first stage are used to condition the second stage along with the caption embedding to generate higher resolution ($256 \times 256$) images, with significant improvement in quality. Conditional augmentation was introduced to ensure continuity on the latent space of text embedding while maintaining the same richness in the newly induced text space. This is ensured by sampling from a Gaussian distribution whose mean vector and covariance matrix is a function of the caption. Most recently, to be able to consider appropriate parts of a given text description, AttnGAN \cite{xu2017attngan} makes use of an attention mechanism, along with a multi-stage GAN, for improved image generation. A hierarchical nested network is proposed in \cite{zhang2018photographic} to assist the generator in capturing complex image statistics.

Despite the aforementioned few efforts, it is worth noting that all the methods so far in the literature use only one caption to generate images, despite the availability of multiple captions in datasets today. Our proposed method iteratively improves the image quality by distilling concepts from multiple captions. Our extensive experimentation stands a testimony to the claim that utilization of multiple captions isn't merely an aggregation of object mentions, but a harmony of complex structure of objects and their relative relations. To strengthen our claim, we quote one such work \cite{sharma2018chatpainter} that loosely corresponds to our idea. The authors improve the image quality by taking into account the dialogues (questions and answers) about an image along with the captions. Though the work shows impressive improvement, the process of answer collection is not similar to multi-captioning and imposes an extra overhead to the system thereby, aggravating the supervision budget for intricate images. This separates our work from their effort. 
\begin{figure*}[h]
\centering
\subfloat[$Legend: \{\textbf{t}_i\}:\text{True Captions; }\{\hat{\textbf{t}}_i\}:\text{Generated Captions; } \{\hat{\textbf{I}}_i\}:\text{Generated Images; } \textbf{I}:\text{True Image; } \newline \text{CCCL: \ccc Loss;}
\text{ DL: Discriminator Loss.}$]{\includegraphics[width=0.75\linewidth]{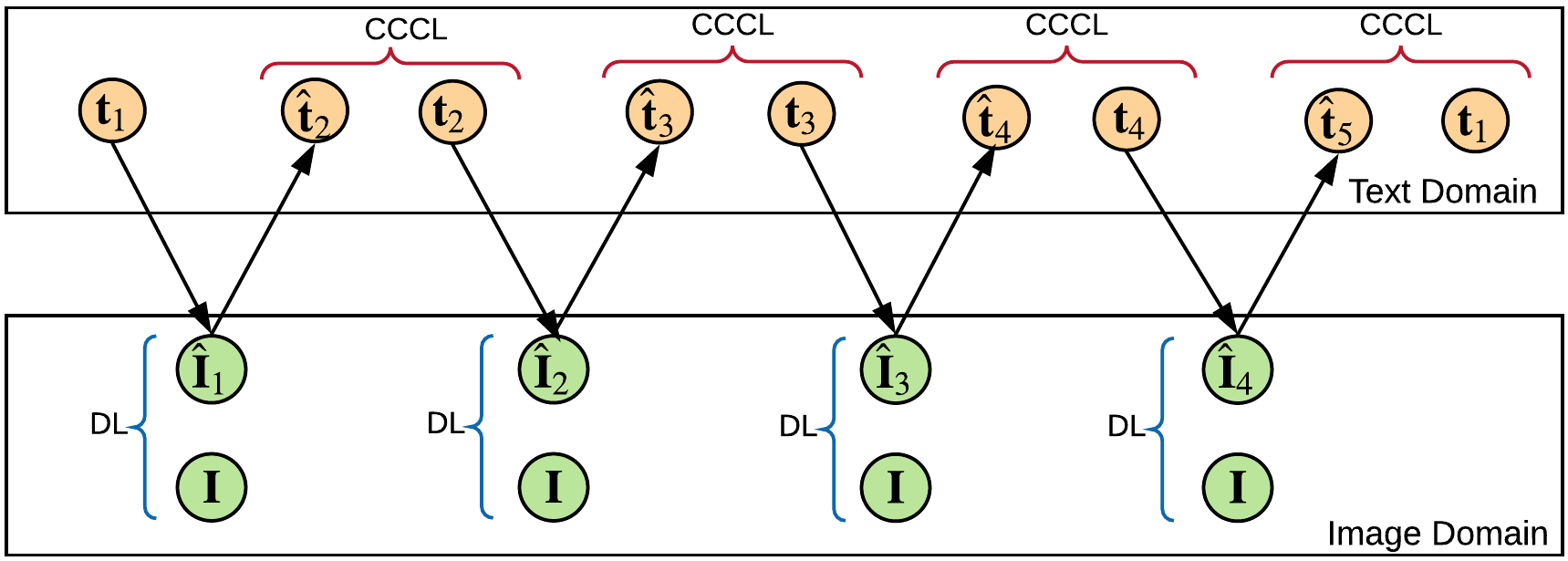}}\qquad
\caption{The figure shows how \ccc is maintained across four captions $(\textbf{t}_1, \cdots, \textbf{t}_4)$. A generator $G$ converts $\textbf{t}_i$ to an image $\hat{\textbf{I}}_i$. Discriminator at each  step forces $\hat{\textbf{I}}_i$ to be realistic. A \ccc Network (CCCN) converts $\hat{\textbf{I}}_i$ back to a caption $(\hat{\textbf{t}}_{i+1})$. The Cross Caption Consistency Loss (CCCL) forces it to be close to $\textbf{t}_{i+1}$. In the last step, $\hat{\textbf{t}}_{5}$ is ensured to be consistent with the initial caption $\textbf{t}_{1}$, hence completing a cycle. 
}
\label{fig:cycleConsistency}
\end{figure*}

\textbf{Cycle Consistency:} Cycle-consistent adversarial networks, i.e. CycleGAN \cite{zhu2017unpaired}, has shown impressive results in unpaired image-to-image translation. CycleGAN learns two mappings, $G:A \to B$ and $F:B \to A$ using two generators $G$ and $F$. $A$ and $B$ can be unpaired images from any two domains. For learning the mapping, they introduce a cycle-consistency loss that checks if $F(G(A))\approx A$ and $G(F(B))\approx B$. Standard discriminator loss ensures that the images generated by $G$ and $F$ are plausible. Several methods like \cite{yi2017dualgan, zhang2018translating, liu2017unsupervised, kim2017learning} with similar ideas has extended the CycleGAN idea more recently in literature. All of them consider only pairwise cycle consistency to accomplish real-world applications such as sketch-to-image generation, real image-to-anime character generation, etc. Our proposed approach takes the idea one step ahead and imposes a transitive consistency across multiple captions. We call this \ccc, which is explained in Section \ref{sec:cycleConsistency}.

\textbf{Recurrent GAN Architectures:} Recurrent GAN was proposed to model data with a temporal component. In particular, Vondrick \etal \cite{DBLP:journals/corr/VondrickPT16}  uses this idea to generate small realistic video clips and, Ghosh \etal \cite{ghosh2016contextual} depict the use of a Recurrent GAN architecture to make predictions on abstract reasoning tasks by conditioning on the previous context or the history. More relevant examples come from the efforts in \cite{DBLP:journals/corr/ImKJM16,benmalek2018neural}, which display the potential of recurrent GAN architectures in generating better quality images. The generative process spans across time, building the image step by step. \cite{benmalek2018neural} utilizes this time lapse to enhance an attribute of the object at a time. We exploit this recurrent nature to continually improve upon the history while generating an image. Unlike previous efforts, we differ in how we model and use the recurrent aspect of the model, and how we update the hidden state of the recurrent model. To the best of our knowledge, the proposed architecture is the first to use a recurrent formulation for the text-to-image synthesis problem.
\begin{figure*}[h]
\centering
{\includegraphics[width=0.99\linewidth]{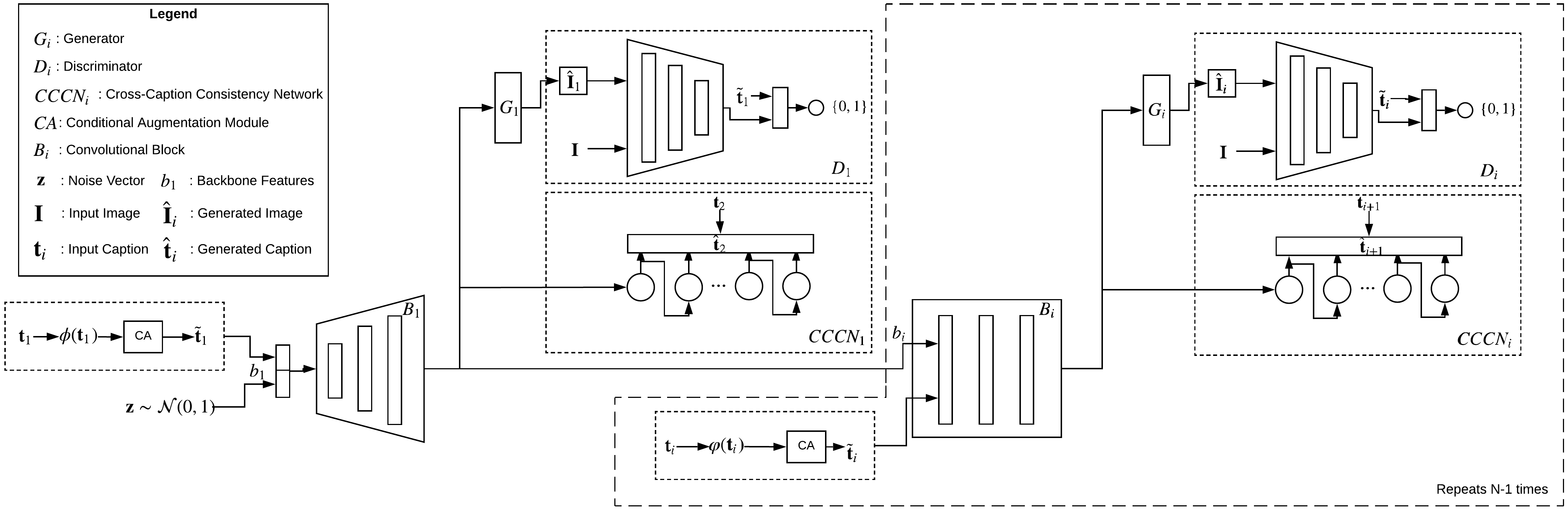}}\qquad
\caption{Figure depicts the cascaded architecture of \first. A series of generators conditioned on $N$ captions one by one and previously generated image through a non-linear mapping (convolutional block $B_i$).  Presently, the value $N$ is set to be 3.}
\label{fig:architecture}
\end{figure*}
\section{Preliminaries} 
\subsection{Generative Adversarial Networks}
GANs are generative models that sidestep the difficulty in approximating intractable probabilistic computations associated with maximum likelihood estimation and related strategies by matching the generative model ($G$) with an adversary ($D$), which learns to discriminate whether the samples are coming from the model distribution ($p_{data}(\textbf{x})$) or the data distribution ($p_\textbf{z}(\textbf{z})$). $G$ and $D$ play the following min-max game with the value function $V(D, G)$:
\begin{align*}
\min_G \max_D V(D, G)&=
\mathbb{E}_{\textbf{x}\sim p_{data}(\textbf{x})}[\log D(\textbf{x})] \\
&+ \mathbb{E}_{\textbf{z}\sim p_\textbf{z}(\textbf{z})}[\log(1 - D(G(\textbf{z})))] 
\end{align*}

In the proposed architecture, we make use of a conditional GAN \cite{mirza2014conditional} in which both the generator and the discriminator are conditioned on a variable $\phi(\textbf{t})$ yielding $G(\textbf{z}| \phi(\textbf{t}))$ and $D(\textbf{x}| \phi(\textbf{t}))$, where $\phi(\textbf{t})$ is a vector representation of the caption.

\subsection{Text embedding}\label{sec:textEmbedding}
The text embedding of the caption that we use to condition the GAN, would yield best results if it could bear a semantic correspondence with the actual image that it represents. One method for such a text encoding is Structured Joint Embeddings (SJE) initially proposed by Akata \etal \cite{akata2015evaluation} and further improved by Reed \etal \cite{reed2016learning}.
They learn a text encoder, $\varphi(\textbf{t})$,  which transforms the caption $\textbf{t}$ in such a way that its inner product with the corresponding image embedding, $\theta(v)$, will be higher if they belong to the same class and lower otherwise. For a training set $\{(v_n, t_n, y_n: n = 1, \cdots, N)\}$, where $v_n$, $t_n$ and $y_n$ corresponds to image, text and the class label, $\varphi(\textbf{t})$ is learned by optimizing the following structured loss:
\vspace{-0.18cm}
\[\frac{1}{N} \sum_{n = 1}^{N} \Delta(y_n, f_v(v_n)) + \Delta(y_n, f_t(t_n))  
\text{ where} \]
\vspace{-10px}
\begin{align*}
f_v(v) &= \operatorname*{arg\,max}_{y \in Y} \mathbb{E}_{t\sim T(y)}[\theta(v)^T\varphi(t)]
\text{ and } \\
f_t(t) &= \operatorname*{arg\,max}_{y \in Y} \mathbb{E}_{v\sim V(y)}[\theta(v)^T\varphi(t)]
\end{align*}
After the network is trained \cite{akata2015evaluation}, we use $\varphi(\textbf{t})$ to encode the captions. Similar method has been used in previous methods for text to image generation \cite{reed2016generative, han2017stackgan, Han17stackgan2, sharma2018chatpainter, Tao18attngan}. $\varphi(\textbf{t})$ is a high dimensional vector. To transform it to a lower dimensional conditioning latent variable, Han \etal \cite{han2017stackgan} proposed the `Conditional Augmentation' technique. Here, the latent vector is randomly sampled from an independent Gaussian distribution whose mean vector and covariance matrix is parameterized by $\varphi(\textbf{t})$. We request the reader to refer to \cite{han2017stackgan} for more information.

\section{Methodology}
\label{sec_methodology}
The main contribution of our work is to formulate a framework to generate images by utilizing information from multiple captions. This is achieved by ensuring \ccc. The generic idea of \ccc is explained in Section \ref{sec:cycleConsistency}. 
We devise two network architectures  that maintain this consistency. The first one is a straightforward instantiation of the idea, where multiple generators progressively generate images by 
consuming captions one by one. This method is explained in Section \ref{sec:first-architecture}. 
A serious limitation of this approach is that the network architecture restricts the number of captions that can be used to generate an image. This leads us to formulate a recurrent version of the method, where a single generator recursively consumes any number of captions. This elegant method is explained in Section \ref{sec:RecurrentMethod}.

\subsection{\ccc} \label{sec:cycleConsistency}
\ccc is achieved by ensuring that the generated image is consistent with a set of captions describing the same image. 
Figure \ref{fig:cycleConsistency} gives a simplified overview of the process. Let us take an example of synthesizing an image by distilling information from four captions. In the first iteration, a generator network ($G$) takes noise and the first caption, $\textbf{t}_1$, as its input, to generate an image, $\hat{\textbf{I}}_1$, which is passed to the discriminator network ($D$), which verifies whether it is real or not. As in a usual GAN setup, generator tries to create better looking images so that it can fool the discriminator. The generated image features are passed on to a `\ccc Network' (CCCN) which will learn to generate a caption for the image. While training, the \ccc Loss ensures that the generated caption is similar to the second caption, $\textbf{t}_2$. 

In the next iteration, $\hat{\textbf{I}}_1$ and $\textbf{t}_2$ is fed to the generator to generate $\hat{\textbf{I}}_2$. While $D$ urges $G$ to make $\hat{\textbf{I}}_2$ similar to the real image $\textbf{I}$, the CCCN ensures that the learned image representation is consistent for generating the next caption in sequence. This repeats until when $\hat{\textbf{I}}_4$ gets generated. Here, the CCCN will ensure that the generated caption is similar to the first caption, $\textbf{t}_1$. Hence we complete a cycle: $\textbf{t}_1 \to \textbf{t}_2 \to \textbf{t}_3 \to \textbf{t}_4 \to \textbf{t}_1$, while generating $\hat{\textbf{I}}_1 \cdots \hat{\textbf{I}}_4$ in-between. $\hat{\textbf{I}}_4$ contains the concepts from all the captions and hence is much richer in quality.

\begin{figure*}[h]
\centering
{\includegraphics[width=0.9\linewidth, height=0.35\linewidth]{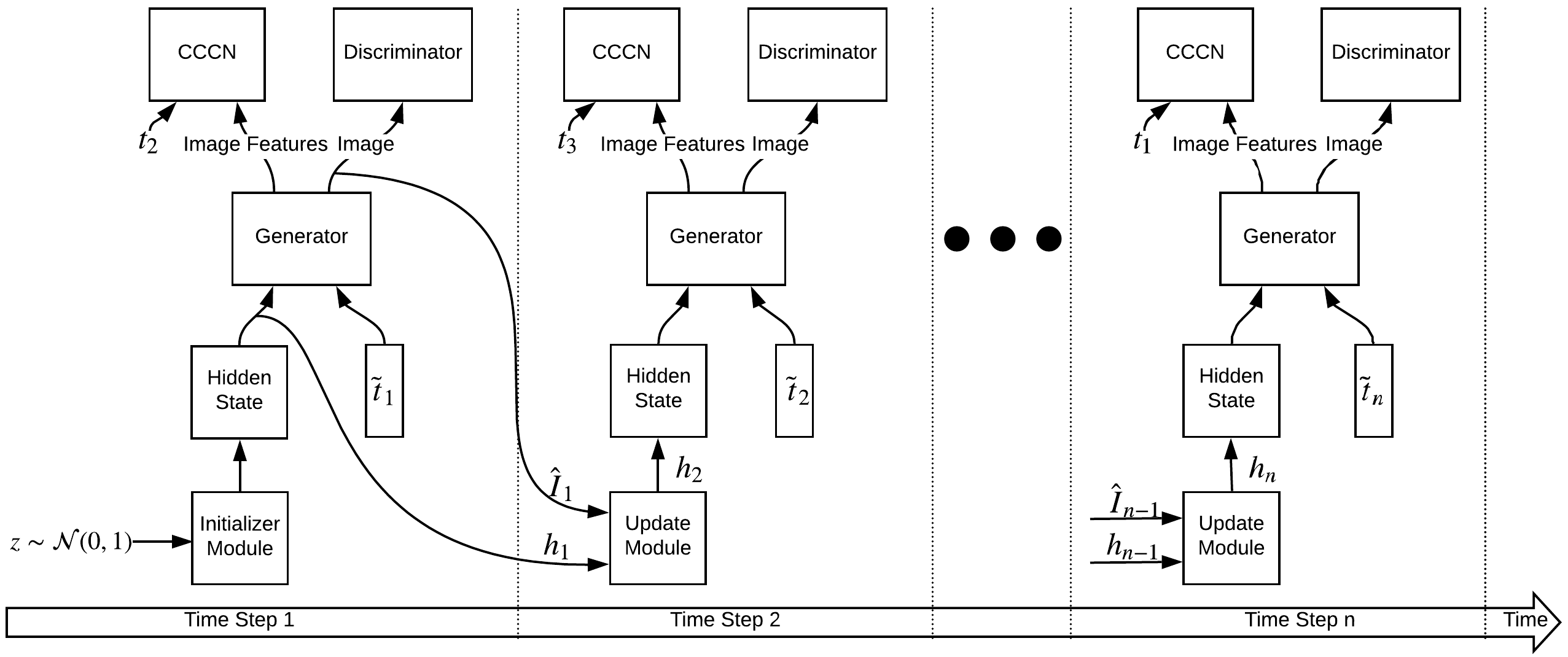}}\qquad
\caption{Architecture of \second. The figure shows the network unrolled in time. $h_i$ refers to the hidden state at time step $i$. $t_i$ is the caption and $\hat{t}_i$ is the vector representation of $t_i$ at time step $i$.}
\label{fig:recurrent_architecture}
\end{figure*}

\subsection{\first} \label{sec:first-architecture}
In our first approach, we consider Cross-caption Cycle Consistent image generation as a cascaded process where a series of generators consumes multiple captions one by one, to generate images. The image that is generated at each step is a function of the previous image and the caption supplied at the current stage. This enables each stage to build up on the intermediate images generated in the previous stage, by utilizing the additional concepts from the new captions seen in the current stage. 
At each stage, a separate discriminator and CCCN is used. 
The discriminator is tasked to identify whether the generated image is fake or real while the CCCN translates the image to its corresponding caption and checks how much close it is to the next caption in succession.

The architecture is presented in Figure \ref{fig:architecture}. 
A set of convolutional blocks (denoted by $B_i$, in the figure) builds up the backbone of the network. The first layer of each $B_i$ consumes a caption. Each generator ($G_i$) and CCCN ($CCCN_i$) branches off from the last layer of each $B_i$, while a new $B_i$ attaches itself to grow the backbone. The number of $B_i$'s is fixed while designing the architecture and restricts the number of captions that can be used to generate an image. The main components of the architecture are explained below.

\subsubsection{Backbone Network} \label{sec:backbone}
\vspace{-0.18cm}
A vector representation ($\tilde{t}_i$) for each caption ($t_i$) is generated using Structured Joint Embedding ($\phi(t_i)$) followed by Conditional Augmentation module. A brief description of the text encoding is presented in Section \ref{sec:textEmbedding}. $\tilde{t}_i$ is a vector of 128 dimension. 
In the first convolutional block, $B_1$, $\tilde{t}_1$ is combined with a 100 dimensional noise vector ($z$), sampled from a standard normal distribution. The combined vector is passed through fully connected layers and then reshaped into $4 \times 4 \times 16 N_g$  tensor. Four up-sampling layers, up-samples the tensor to $64 \times 64 \times 8 N_g$  tensor. This tensor is passed on to the first generator ($G_1$) and the first CCCN ($CCCN_1$). 

Further convolutional blocks, $B_i$, are added to $B_1$ as follows. The new caption encoding $\tilde{t}_i$, is spatially replicated at each location of the backbone features ($b_i$) coming from the previous convolutional block ($B_i$), followed by a $3 \times 3$ convolution. These features are passed thorough a residual block and an up-sampling layer. This helps to increase the spatial resolution of the feature maps in each $B_i$. Hence, the output of $B_2$ is a tensor of size $128 \times 128 \times 4 N_g$. The generator and CCCN branches off from this feature map as before, and a new convolutional block ($B_3$) gets added. In our experiments, we used three $B_i$s, due to GPU memory limitations. $N_g$ is set to $32$.
\subsubsection{Generator}
\vspace{-0.18cm}
Each generator ($G_i$) takes the features from the backbone network and passes it through a single $3\times3$ convolutional layer followed by a tanh activation function to generate an RGB image. As the spatial resolution of the features from each $B_i$ increases (as explained in Section \ref{sec:backbone}), the size of the image generated by each generator, also  increases.  

Multiple generators are trained together by minimizing the following loss function:
\vspace{-10pt}
\begin{align*}
\mathcal{L}_{G}= \sum_{i = 1}^N \mathcal{L}_{G_i} &\text{, where } \mathcal{L}_{G_i} = \mathbb{E}_{\textbf{s}_i\sim p_{G_i}}[\log(1 - D_i(\textbf{s}_i))] \\
\vspace{-1cm}
&+ \lambda D_{KL}(\mathcal{N}(\mu(\phi(\textbf{t}_i), \Sigma(\textbf{t}_i)) || \mathcal{N}(0,1))
\end{align*}
The first term in $\mathcal{L}_{G_i}$ is the standard minimization term in the GAN framework which pushes the generator to generate better quality images. $p_{G_i}$ is the distribution of the generator network.
The $D_{KL}$ term is used to learn the parameters of $\mu(\phi(\textbf{t}_i))$ and $\Sigma(\textbf{t}_i)$ of the Conditional Augmentation framework \cite{han2017stackgan}. It is learned very similar to the re-parameterization trick in VAEs \cite{kingma2013auto}. $\lambda$ is a regularization parameter, whose value we set to 1 for the experiments.
\subsubsection{Discriminator}
\vspace{-0.18cm}
The discriminators ($d_i$) contains a set of down-sampling layers which converts the input tensor to $4\times 4\times 18N_d$ tensor. Following the spirit of conditional GAN \cite{mirza2014conditional}, the encoded caption, $\tilde{t}_i$ is spatially replicated and joined by a $3\times3$ convolution to the incoming image. The final logit is obtained by convolving with a $4\times4\times1$ kernel and a Sigmoid activation function. 

The loss function for training the discriminator is as follows:
\vspace{-0.18cm}
\[\mathcal{L}_{D_i}=
\mathbb{E}_{\textbf{I}_i\sim p_{data}}[\log D_i(\textbf{I}_i)]
+ \mathbb{E}_{\textbf{s}_i\sim p_{G_i}}[\log(1 - D_i(\textbf{s}_i))]\]
$p_{data}$ is the original data distribution and $p_{G_i}$ is the distribution of the corresponding generator network. The multiple discriminators are trained in parallel.
\vspace{-0.25cm}
\subsubsection{\ccc Network} \label{sec:cccn_arch}
\vspace{-0.18cm}
CCCN is modeled as an LSTM which generates one word at each time-step conditioned on a context vector (derived by attending to specific regions of the image), the hidden state and the previously generated word. 
CCCN takes as input the same set of backbone features that the generator consumes. It is then pooled to reduce the spatial dimension. Regions of these feature maps are aggregated into a single context vector by learning to attend to these feature maps similar to the method proposed by \cite{xu2015show}.
Each word is encoded as its one-hot representation. 

There is one CCCN block per generator. CCCN is trained by minimizing the cross-entropy loss between each of the generated words and words in the true caption. The true caption for Stage $i$ is $(i+1)^{th}$ caption, and finally the first caption, as is explained in Section \ref{sec:cycleConsistency}. The loss of each of the CCCN block is aggregated and back-propagated together. 

\subsection{\second} \label{sec:RecurrentMethod}
\vspace{-0.15cm}
The architecture of \first limits the number of captions that can be consumed because the number of generator-discriminator pairs has to be decided and fixed during training. We overcome this problem by formulating a recurrent approach for text to image synthesis. At its core, \second maintains a hidden state, which guides the image generation at each time step, along with a new caption. The hidden state by itself is modeled as a function of the previous hidden state and the image that was generated in the previous time step. This allows the hidden state to act like a shared memory, that captures the essential features from the different captions to generate good looking, semantically rich images. The exact way in which hidden state is updated is explained in Section \ref{sec:Updating_hidden_state}.

Figure \ref{fig:recurrent_architecture} presents the simplified architecture of \second. We explain the architecture in  detail here. The hidden state is realized as an $8 \times 8 \times 8$ tensor. The values for the initial hidden state is learned by the Initializer Module, which takes as input a noise vector ($z$) of length 100, sampled randomly from a Gaussian distribution. It is passed though a fully connected layer followed by non linearity and finally reshaped into a $8 \times 8 \times 8$ tensor. Our experimentations reveal that initializing the hidden state with Initializer Module helps the model to learn better than randomly initializing the same.

The hidden state along with the text embedding of the caption is passed to the generator to generate an Image. A discriminator guides the generator to generate realistic image while a \ccc Network (CCCN) ensure that the captions that are generated from the image features are consistent with the second caption. As we unroll the network in time, different captions are fed to the generator at each time step. When the final caption is fed in, the CCCN makes sure that it is consistent with the first caption. Hence the network ensures that the cycle consistency between captions is maintained. 

The network architecture of CCCN is same as that of \first, while the architecture of the generator and discriminator is slightly different. We explain them in section \ref{sec:second_gen_disc}. While \first has separate generator, and the corresponding discriminator and CCCN at each stage, the \second has only one generator, discriminator and CCCN. The weights of the generator is shared across time steps and is updated via Back Propagation Through Time (BPTT)\cite{williams1995gradient}. 

\begin{figure*}
\centering
{\includegraphics[scale=0.144]{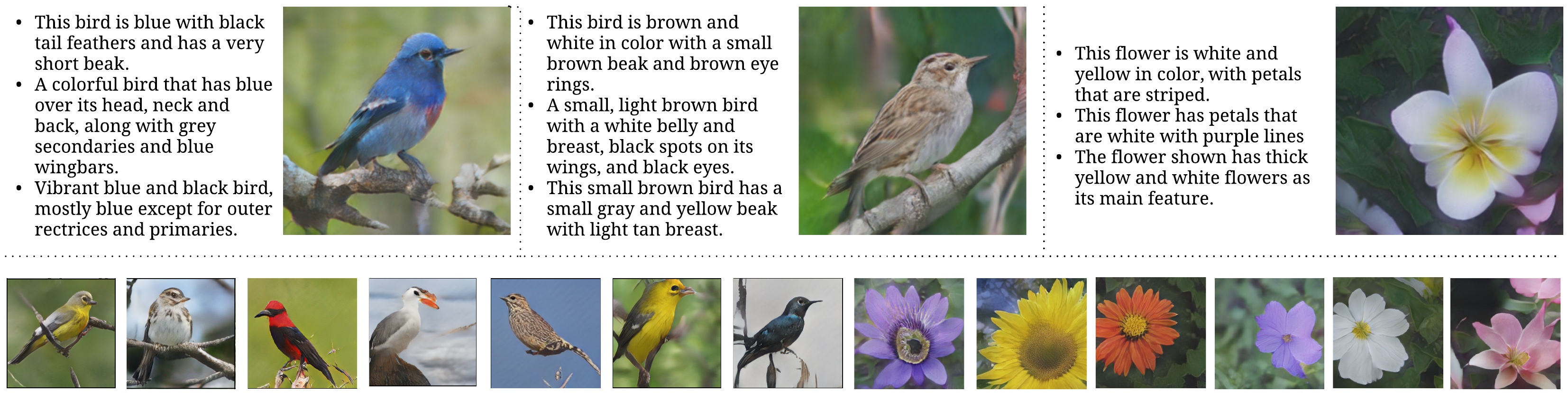}}\qquad
\caption{Generations from \first. The first row shows the images generated and the corresponding captions consumed in the process. The first two images belong to Indigo Bunting, Tree Sparrow class of CUB dataset \cite{WelinderEtal2010} and the last image belongs to Peruvian Lily class  of Flowers dataset \cite{Nilsback08}. 
The bottom row showcases some random samples of generated images. 
(Kindly zoom in to see the detailing in the images.)}
\label{fig:results_cascade}
\end{figure*}

\begin{figure*}
\centering
{\includegraphics[scale=0.15]{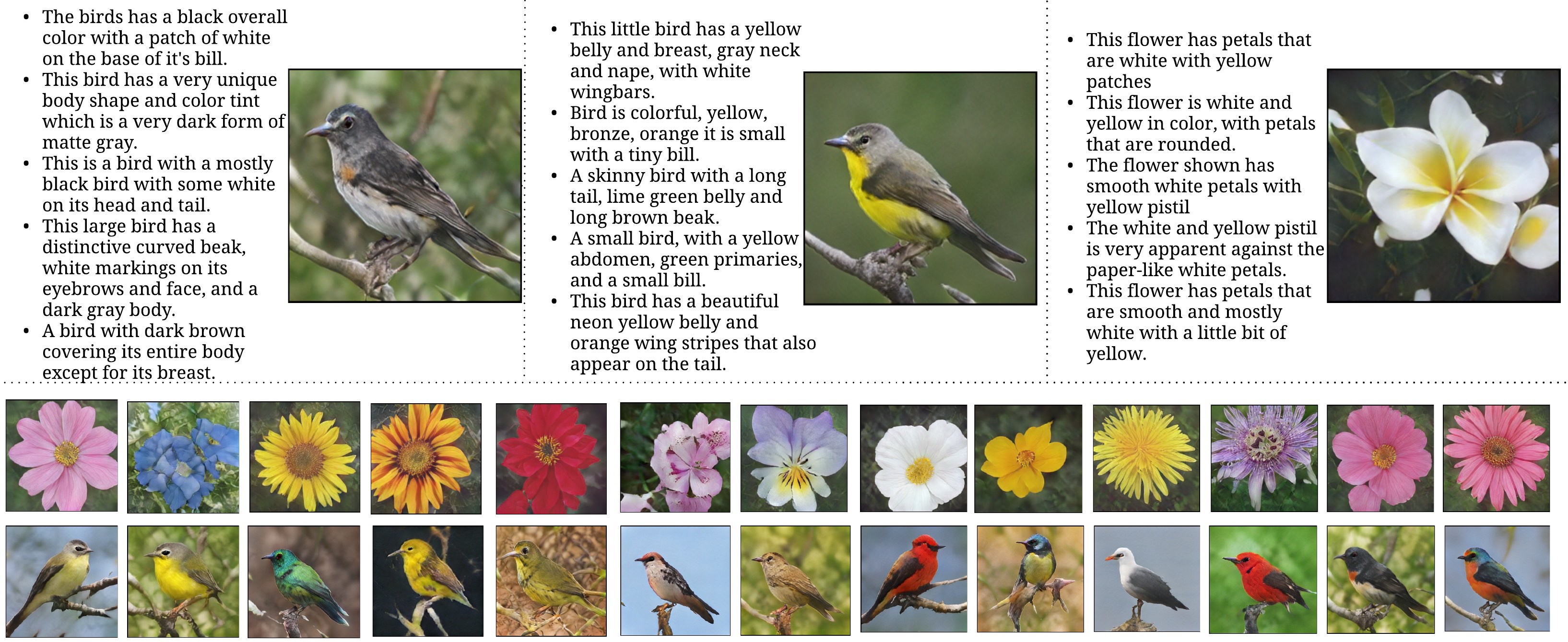}}\qquad
\caption{Generations from \second. The first two images are generated from the caption belonging to Black Footed Albatross class and Great Crested Flycatcher class of CUB dataset \cite{WelinderEtal2010}, while the last one is from the Moon Orchid class of Flowers dataset \cite{Nilsback08}. The last two rows contains random generations from both the datasets. (Kindly zoom in to see the detailing in the images.)}
\label{fig:results_recurrent}
\end{figure*}

\subsubsection{Updating the Hidden State} \label{sec:Updating_hidden_state}
In the first time step of the unrolled network, the hidden state is initialized by the Initializer Module. In the successive time steps, the hidden state and the image generated in the previous time step is used to generate the new hidden state, as shown in figure \ref{fig:recurrent_architecture}. The $64 \times 64$ images are down-sampled by a set of down-sampling convolutional layers to generate feature maps of spatial dimension $8 \times 8$. These feature maps are fused with the hidden state (also of spatial dimension $8\times8$) by  eight $3\times3$ filters. This will result in a new hidden state of dimension $8\times8\times8$. If we denote the above operation by a function $UpdateHiddenState(.)$, then the recurrence relation at each time-step i, can be expressed as:
\begin{align*}
\hat{I}_i &= Generator(h_i, \phi(t_1)) \\
h_i &= UpdateHiddenState(h_{i-1}, I_{i-1})
\end{align*}
$\hat{I}_i$ is the image generated by the Generator, by consuming the  hidden state ($h_i$) and the vector representation of the caption ($\phi(t_1)$) that was provided in time step $i$.

\subsubsection{Generator and Discriminator} \label{sec:second_gen_disc}
\second uses a single generator to generate images of size $256\times256$. It consumes the hidden state $h_i$,  and a vector representation $\phi(t_i)$ of the caption provided in the current time step. $\phi(t_i)$ is spatially replicated  to each location of the hidden state and then fused by a $3\times3$ convolution layer. This results in a feature map of spatial resolution $8\times8$.

One easy way to generate $256\times256$ images from these feature maps would be to stack five up-convolution layers (each doubling the spatial resolution) back to back.
Our experiments showed that such a method will not work in practice. Hence, we choose to generate intermediate images of spatial resolution $64\times64$ and $128\times128$ also. This is achieved by attaching $3\times3\times3$ kernels after the third and fourth up-sampling layer. The extra gradients (obtained by discriminating the intermediate images) that flow through the network will help the network to learn better. 

In-order to discriminate the two intermediate images and the final image, we make use of three separate discriminators. The architecture of each of the discriminator is similar to \first.

\begin{figure}
\centering
{\includegraphics[scale=0.126]{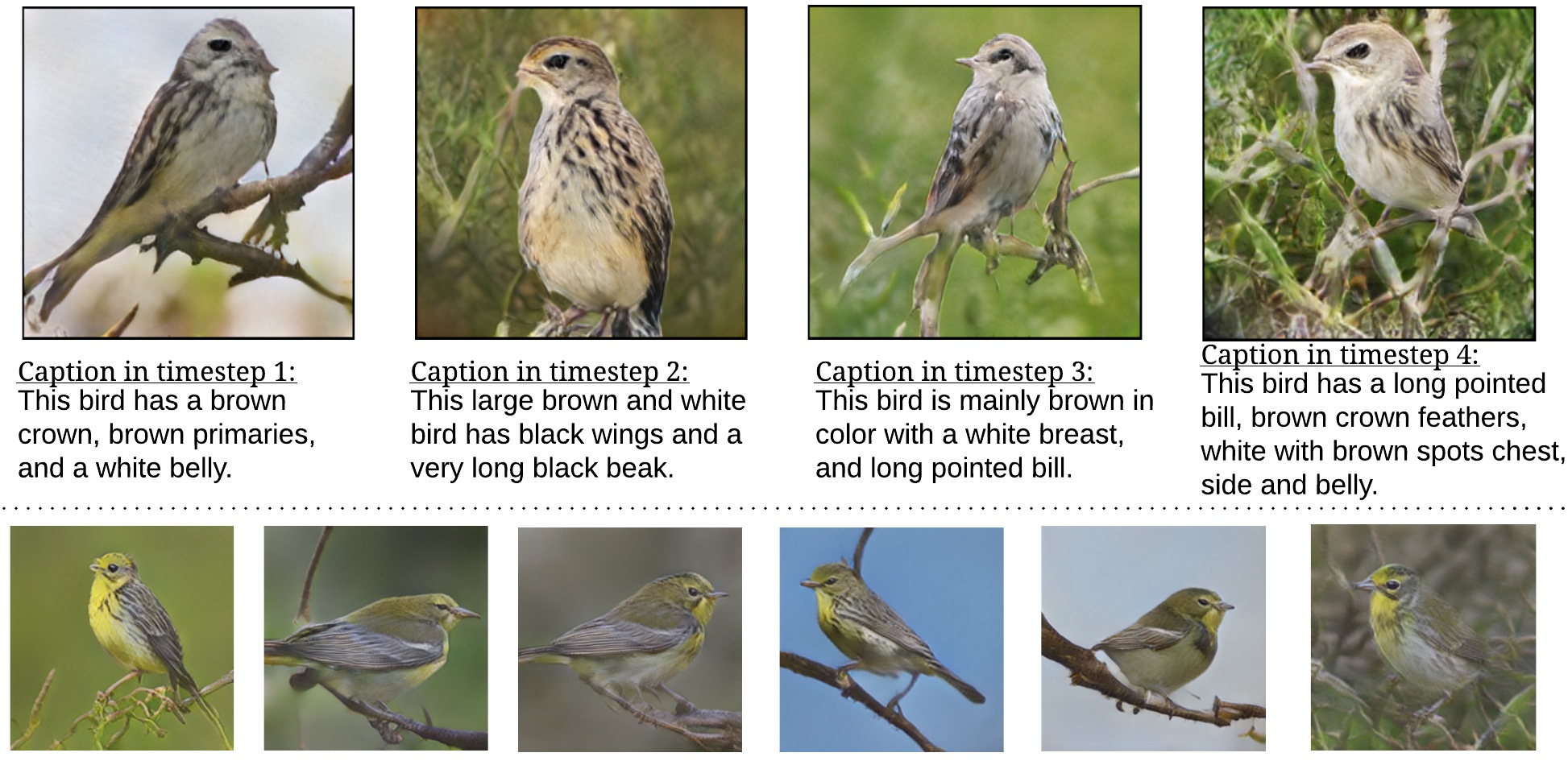}}\qquad
\caption{The top row shows the images generated by \second at each time-step. The corresponding captions that was consumed is also added. The bottom row shows generated birds of the same class, but with varying pose and background. These are generated by keeping the captions the same and varying the noise vector used to condition the GAN.}
\label{fig:result_2}
\end{figure}
\vspace{-0.1cm}
\section{Experiments and Results}\label{sec:results}
\subsection{Datasets and Evaluation Criteria}
We evaluate \first and \second on Oxford-102 flowers dataset \cite{Nilsback08} and Caltech-UCSD Birds (CUB) \cite{WelinderEtal2010} datasets. Oxford-102 contains 102 categories of flowers counting to 8189 images in total, while CUB contains 200 bird species with 11,788 images. Following the previous methods \cite{reed2016generative,han2017stackgan,Han17stackgan2}, we pre-process the dataset to improve the object to image ratio.

We gauge the performance of the generated images by its `Inception Score'\cite{salimans2016improved}, which has emerged as the dominant way of measuring the quality of generative models. The inception model has been fine-tuned on both the the datasets so that we can have a fair comparison with previous methods like \cite{reed2016generative,han2017stackgan,Han17stackgan2,zhang2018photographic}.
\vspace{-0.15cm}
\subsection{Results}
We validate the efficacy of \first and \second by comparing it with GAN-INT-CLS \cite{reed2016generative}, GAWWN \cite{reed2016learning}, StackGAN \cite{han2017stackgan}, StackGAN++ \cite{Han17stackgan2} and HD-GAN \cite{zhang2018photographic} (Our future work will include integrating attention in our framework, and comparing against attention-based frameworks such as \cite{Tao18attngan}). 

\subsubsection{Quantitative Results}
\vspace{-0.14cm}
\begin{table}[h]
\centering
\begin{tabular}{@{}lcc@{}}
\toprule
Method            & Oxford-102 \cite{Nilsback08} & CUB  \cite{WelinderEtal2010} \\ \midrule
GAN-INT-CLS \cite{reed2016generative}      & 2.66 $\pm$ .03 & 2.88 $\pm$ .04    \\
GAWWN   \cite{reed2016learning}          &     -       &   3.62 $\pm$ .07  \\
StackGAN   \cite{han2017stackgan}       & 3.20 $\pm$ .01&  3.70 $\pm$ .04    \\
StackGAN++ \cite{Han17stackgan2}       &     -       &   3.82 $\pm$ .06  \\
HDGAN    \cite{zhang2018photographic}         & 3.45 $\pm$ .07 &  \textbf{4.15 $\pm$ .05}   \\ \midrule
Cascaded C4Synth  & 3.41 $\pm$ .17 &  3.92 $\pm$ .04   \\
Recurrent C4Synth & \textbf{3.52 $\pm$ .15} &   4.07 $\pm$ .13  \\ \bottomrule
\end{tabular}
\caption{Comparison of C4Synth methods with other text to image synthesis methods. The number reported are Inception Scores (higher is better).}
\label{table:result}
\end{table}

Table \ref{table:result} summarizes the Inception Score of competing methods on Oxford-102 flowers dataset \cite{Nilsback08} and Caltech-UCSD Birds (CUB) \cite{WelinderEtal2010} dataset along with the results of C4Synth models. 
On Oxford-102 dataset, \second method gives state-of-the-art result, improving the previous baseline. On CUB dataset, the results are comparable with HDGAN \cite{zhang2018photographic}.

The results indicate that \second has an edge over \first. It is worth noting that both the methods perform better than four out of five other baseline methods.
\vspace{-0.2cm}
\subsubsection{Qualitative results}
Figure \ref{fig:results_cascade} and \ref{fig:results_recurrent} shows the generations from \first and \second methods respectively. The generations from \first method consumes three captions, as is restricted by the architecture, while the \second method consumes five captions. The quality of the images generated by both the methods are comparable as is evident from the Inception Scores.
All the generated images are of $256 \times 256$ pixels in resolution. The supplementary section contains more image generations. 

The images that are generated at each time step by the \second method is captured in the top row of Figure \ref{fig:result_2}. The captions that are consumed in each step is also shown. This figure validates our assertion that the recurrent formulation progressively generates better images by consuming one caption at a time.

The bottom row of Figure \ref{fig:result_2} shows the interpolation of the noise vector, used to generate the hidden state of \second, while fixing the captions used. This results in generating the same bird in different orientations and backgrounds.

\vspace{-5pt}
\subsubsection{Zero Shot generations}
We note that while training both the \method architectures with  Oxford-102 flowers dataset \cite{Nilsback08} and Caltech-UCSD Birds (CUB) \cite{WelinderEtal2010} datasets, the classes used for training and testing are disjoint. 
We use the official train-test split for both the datasets. CUB has 150 train+val classes and 50 test classes, while Oxford-102 has 82 train+val classes and 20 test classes. 
Hence all the results shown in the paper are zero-shot generations, where none of the classes of captions that are used to generate the image in test phase, has ever been seen in training phase. 
\vspace{-5pt}
\section{Conclusion}
We formulate two generative models for text to image synthesis, \first and \second, which makes use of multiple captions to generate an image. The method is able to generate plausible images on Oxford-102 flowers dataset \cite{Nilsback08} and Caltech-UCSD Birds (CUB) \cite{WelinderEtal2010} dataset.
We believe that attending to specific parts of the captions at each stage, would improve the results of our method. We will explore this in a future work. The code is open-sourced at http://josephkj.in/projects/C4Synth.

{\small
\bibliographystyle{ieee}
\bibliography{egbib}
}

\end{document}